\documentclass{article}
\usepackage{spconf,amsmath,graphicx,hyperref}
\usepackage{subcaption}

\usepackage{algorithm,amsmath,algpseudocode}
\usepackage{multirow}
\usepackage{comment}

\usepackage{fancyhdr}

\fancypagestyle{firstpage}{
  \fancyhf{} 
  \fancyfoot[C]{\footnotesize © 20XX IEEE. Personal use of this material is permitted. Permission from IEEE must be obtained for all other uses, in any current or future media, including reprinting/republishing this material for advertising or promotional purposes, creating new collective works, for resale or redistribution to servers or lists, or reuse of any copyrighted component of this work in other works.} 
}

\title{CONTROLLABLE LOCALIZED FACE ANONYMIZATION VIA DIFFUSION INPAINTING}
\name{Ali Salar$^{2,1}$, Qing Liu$^{3,1}$, Guoying Zhao$^{*2,1}$}
\address{{\small 1) Center for Machine Vision and Signal Analysis (CMVS), University of Oulu, Finland - 2) ELLIS Institute Finland}\\ \small 3) The Machine Learning Group, UiT The Arctic University of Norway, Norway \\ \quad
{\tt\small \{ali.salar, qing.liu, guoying.zhao\}@oulu.fi}}

\begin{document}

\maketitle
\let\thefootnote\relax\footnotetext{$^{*}$ indicates the corresponding author}
\let\thefootnote\relax\footnotetext{$^{1}$ \href{https://github.com/parham1998/Face-Anonymization}{https://github.com/parham1998/Face-Anonymization}}

\begin{abstract}
The growing use of portrait images in computer vision highlights the need to protect personal identities. At the same time, anonymized images must remain useful for downstream computer vision tasks. In this work, we propose a unified framework that leverages the inpainting ability of latent diffusion models to generate realistic anonymized images. Unlike prior approaches, we have complete control over the anonymization process by designing an adaptive attribute-guidance module that applies gradient correction during the reverse denoising process, aligning the facial attributes of the generated image with those of the synthesized target image. Our framework also supports localized anonymization, allowing users to specify which facial regions are left unchanged. Extensive experiments conducted on the public CelebA-HQ and FFHQ datasets show that our method outperforms state-of-the-art approaches while requiring no additional model training. The source code is available on our page$^{1}$.
\end{abstract}

\section{Introduction}
\label{sec:intro}
\thispagestyle{firstpage}

With the rise of social media sharing and powerful recognition algorithms, facial images have become highly sensitive personal data. Regulations such as the General Data Protection Regulation (GDPR) \cite{GDPR} mandate robust privacy protection, yet they also pose challenges for research fields that rely on facial datasets. This highlights the need for anonymization methods that protect identity while maintaining data utility for downstream computer vision tasks, such as healthcare \cite{health_ToAC_2021}.\par

\begin{figure}[t]
    \centering
    \resizebox{0.93\columnwidth}{!}
    {\begin{tabular}{cccc}
        \begin{subfigure}[t]{0.2\textwidth}
            \includegraphics[width=\linewidth]{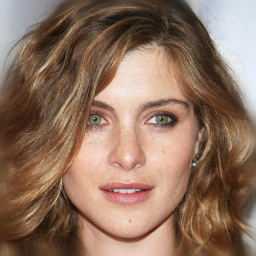}
            \centering
            \vspace{2pt}
            \includegraphics[width=\linewidth]{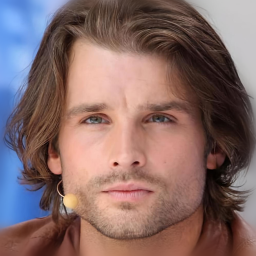}
            \centering
            \vspace{2pt}
            \fontsize{14pt}{16pt}\selectfont{Original Images}
        \end{subfigure}
        \begin{subfigure}[t]{0.2\textwidth}
            \includegraphics[width=\linewidth]{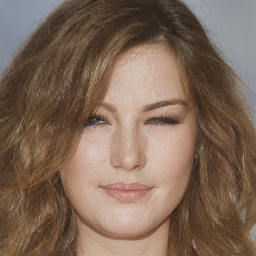}
            \centering
            \vspace{2pt}
            \includegraphics[width=\linewidth]{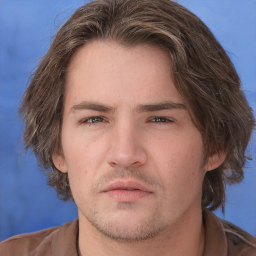}
            \centering
            \vspace{2pt}
            \fontsize{14pt}{16pt}\selectfont{(a) FALCO \\ 2023}
        \end{subfigure}
        \begin{subfigure}[t]{0.2\textwidth}
            \includegraphics[width=\linewidth]{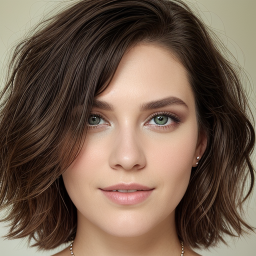}
            \centering
            \vspace{2pt}
            \includegraphics[width=\linewidth]{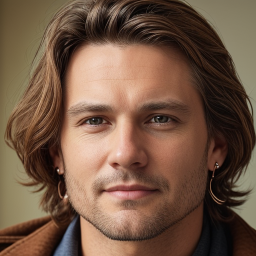}
            \centering
            \vspace{2pt}
            \fontsize{14pt}{16pt}\selectfont{(b)}
            \fontsize{12pt}{14pt}\selectfont{CAMOUFLaGE}
            \fontsize{14pt}{16pt}\selectfont{\\ 2024}
        \end{subfigure}
        \begin{subfigure}[t]{0.2\textwidth}
            \includegraphics[width=\linewidth]{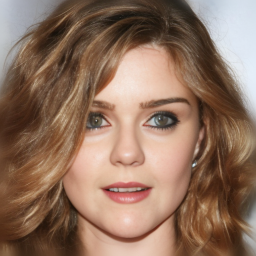}
            \centering
            \vspace{2pt}
            \includegraphics[width=\linewidth]{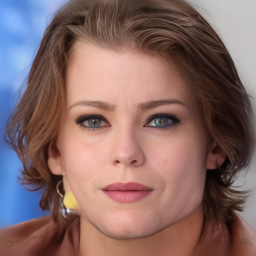}
            \centering
            \vspace{2pt}
            \fontsize{14pt}{16pt}\selectfont{(c) FAMS \\ 2025}
        \end{subfigure}
        \begin{subfigure}[t]{0.2\textwidth}
            \includegraphics[width=\linewidth]{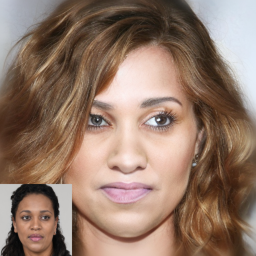}
            \centering
            \vspace{2pt}
            \includegraphics[width=\linewidth]{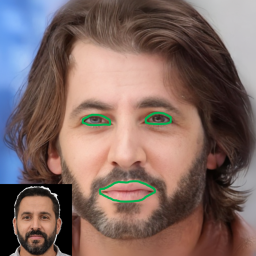}
            \centering
            \vspace{2pt}
            \fontsize{14pt}{16pt}\selectfont{(d) \textbf{Ours}}
        \end{subfigure}
    \end{tabular}}
    \caption{Comparison of our method to FALCO \cite{FALCO_CVPR_2023}, CAMOUFLaGE \cite{CAMOUFLaGE_arXiv_2024}, and FAMS \cite{FAMS_WACV_2025} for identity anonymization. Target images are displayed near the anonymized ones.}
    \label{fig_1}
\end{figure}

Traditional anonymization methods, such as blurring, pixelation, or masking, are simple but significantly degrade image quality and utility, and remain susceptible to reconstruction attacks \cite{Fantômas_2022_arXiv}. Recent Generative Adversarial Network (GAN)-based \cite{GAN_NIPS_2014} approaches generate more realistic results but often fail to preserve non-identifiable attributes \cite{Deepprivacy_visual_computing_2019, CIAGAN_CVPR_2020}. For example, FALCO \cite{FALCO_CVPR_2023} generates completely synthetic images that retain specific facial attributes but alter elements such as background, hairstyle, and accessories, as shown in Fig. \ref{fig_1}(a).\par

More recently, diffusion models \cite{DDPM_2020_ANIPS, DDIM_2021_ICLR, Stable_diff_2022_CVPR} have been explored for face anonymization due to their ability to produce high-quality, realistic images. CAMOUFLaGE \cite{CAMOUFLaGE_arXiv_2024} recreates the entire image to balance privacy and fidelity, preserving facial structures and expression while altering features such as background and hair color (see Fig. \ref{fig_1}(b)). Although it supports adjustable anonymization through a scale parameter, it provides no control over the generated results. Similarly, FAMS \cite{FAMS_WACV_2025} preserves key attributes such as expression, pose, and background, and also allows adjustable anonymization. However, it suffers from two drawbacks: a lack of controllability, sometimes even altering gender, as shown in Fig. \ref{fig_1}(c); and a high computational cost during image generation.\par

Despite these advances, existing methods \cite{FALCO_CVPR_2023, CAMOUFLaGE_arXiv_2024, FAMS_WACV_2025} lack controllable and localized anonymization. Controllable anonymization allows users to modify facial attributes or expressions as desired. For example, generating anonymized faces of the same person with different expressions to support diverse downstream tasks. Localized anonymization, on the other hand, enables selective protection of specific facial regions. This is particularly important in sensitive domains such as medicine, where privacy must be preserved without losing clinically relevant details. For instance, ophthalmologists may require real patient eyes for diagnosis while anonymizing other facial regions.\par

Motivated by these challenges, we propose a novel anonymization framework based on the Stable Diffusion model (SD) \cite{Stable_diff_2022_CVPR}. By leveraging the strong inpainting capabilities of SD, our method restricts edits to the facial region while preserving background, hairstyle, and accessories. We further introduce an adaptive attribute-guidance module to perform gradient correction during the reverse denoising process, reducing the distance between the generated anonymized image and a synthesized target image in the embedding space of the feature-matching model. Since our work differs from face swapping and to avoid ethical concerns, we use a synthesized image as the target image, rather than a real individual's image; hereafter, the target image refers to this synthesized version. This design enables fine control over facial features and expressions in the anonymized output. Moreover, our inpainting strategy supports localized anonymization, allowing users to retain specific regions, such as eyes, lips, or nose, offering greater flexibility to balance privacy with usability (see Fig. \ref{fig_1}(d)). It is important to note that our proposed method requires no additional training, as it directly leverages the inherent properties of pre-trained SD.\par

\section{Methodology}
\label{sec:method}

\subsection{Problem Definition}
Given an original face $x$ and a target image ${x}_{tgt}$, the goal is to generate an anonymized face $x_a$ that conceals the identity of $x$ from face recognition (FR) systems \cite{Facenet_2015_CVPR}, while inheriting controllable attributes from ${x}_{tgt}$ to ensure utility for downstream tasks. The objective can be defined as:
\begin{equation}
    \label{eq1}
    arg\min_{x_{a}} \mathcal{S}\left(\mathcal{F}\left(x\right), \mathcal{F}\left(x_{a}\right)\right)\\
    \text { s.t. } \mathcal{U}(x_a) \approx \mathcal{U}(x_{tgt}),
\end{equation}
where $\mathcal{F}$ denotes the feature extractor used in the FR model, $\mathcal{S}$ represents a similarity measure (e.g., cosine similarity), and $\mathcal{U}$ denotes utility features, such as expression.

\subsection{Proposed Framework}
Fig. \ref{fig_2} shows an overview of the proposed method, which consists of the following main components:

\begin{figure}[h]
    \centering
    \includegraphics[width=\columnwidth]{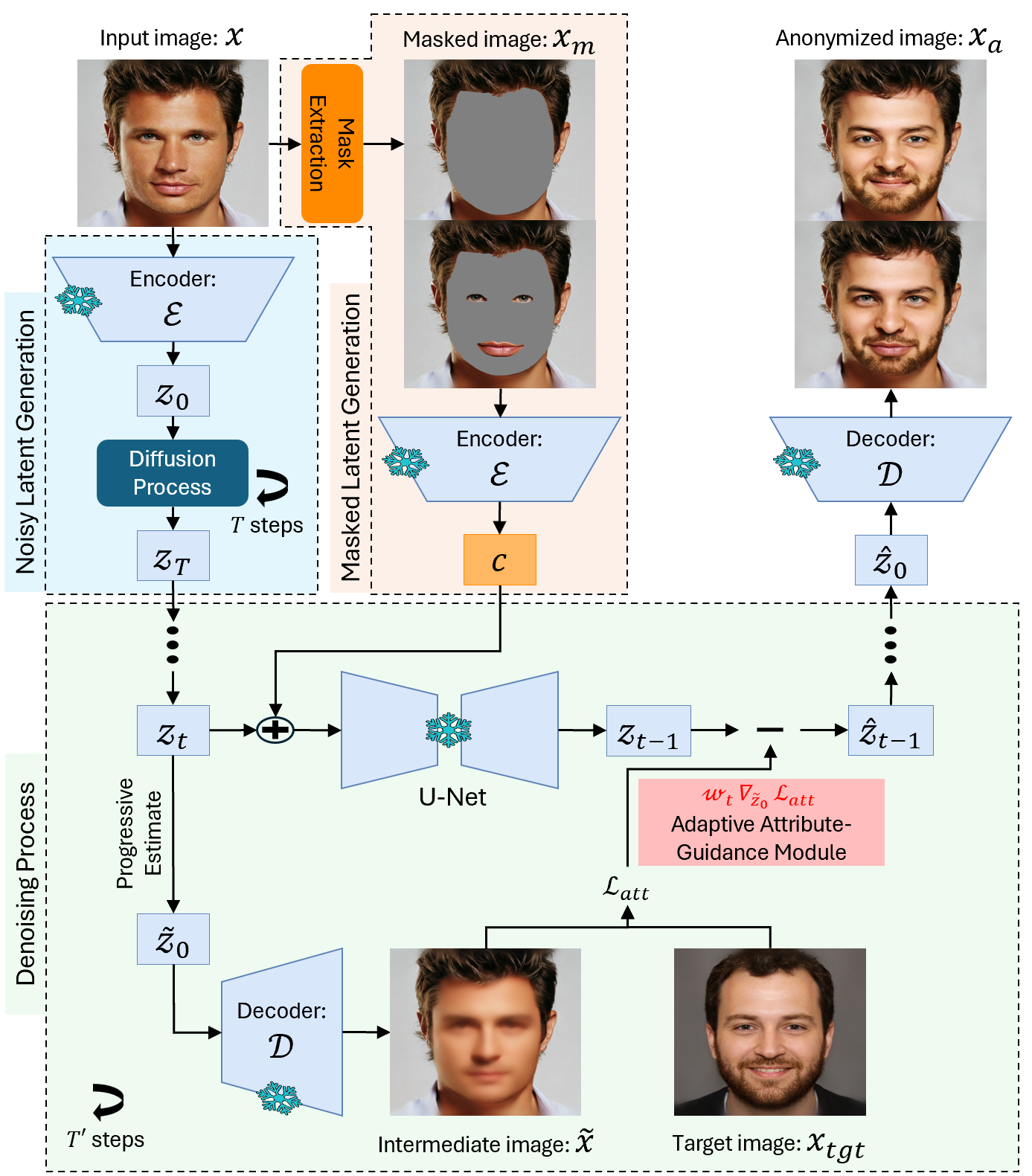}
    \caption{The proposed framework for face anonymization.}
    \label{fig_2}
\end{figure}

\textbf{Noisy Latent Generation.} We leverage the pre-trained encoder $\mathcal{E}$ to encode the input image $x$ into a latent representation $z_0=\mathcal{E}(x)$. A forward diffusion process is then applied to $z_0$ over $T$ steps, producing the noisy latent $z_T$:\par
\begin{equation}
    \label{eq2}
    {z}_T=\sqrt{\bar{\alpha}_T} z_0+\sqrt{1-\bar{\alpha}_T} \epsilon,
\end{equation}
where $\bar{\alpha}_{T}$ is the noise scaling factor and $\epsilon \sim \mathcal{N}(0,1)$.

\textbf{Masked Latent Generation.} To preserve background details and restrict edits to the facial region, we use a masked image as guidance. Specifically, we use a pre-trained face parsing model $f$ \cite{EHANet_MDPI_2020} to obtain a binary mask $M$ where $M_{i,j} \in \{0, 1\}$, and $M_{i,j} = 1$ denotes the pixel $(i,j)$ belongs to the editable facial region, and $M_{i,j} = 0$ denotes the pixel belongs to the preserved background. The masked image $x_m = x \odot (1 - M)$ is then encoded into a latent code $c = \mathcal{E}(x_m)$. which is concatenated with the noisy latent $z_T$ to guide the UNet, ensuring that only the masked region is regenerated.

\begin{figure}[t]
    \centering
    \resizebox{0.8\columnwidth}{!}
    {\begin{tabular}{cccc}
        \begin{subfigure}[t]{0.2\textwidth}
            \includegraphics[width=\linewidth]{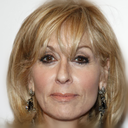}
            \centering
            \vspace{2pt}
            \includegraphics[width=\linewidth]{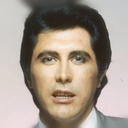}
            \centering
            \vspace{2pt}
            \fontsize{14pt}{16pt}\selectfont{Original Images}
        \end{subfigure}
        \begin{subfigure}[t]{0.2\textwidth}
            \includegraphics[width=\linewidth]{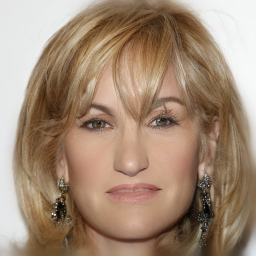}
            \centering
            \vspace{2pt}
            \includegraphics[width=\linewidth]{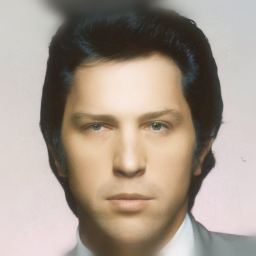}
            \centering
            \vspace{2pt}
            \fontsize{14pt}{16pt}\selectfont{Vanilla}
        \end{subfigure}
        \begin{subfigure}[t]{0.2\textwidth}
            \includegraphics[width=\linewidth]{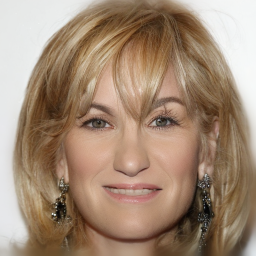}
            \centering
            \vspace{2pt}
            \includegraphics[width=\linewidth]{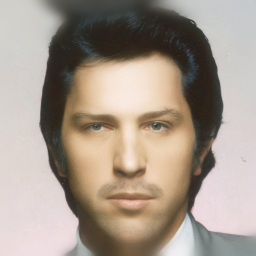}
            \centering
            \vspace{2pt}
            \fontsize{14pt}{16pt}\selectfont{Text guided}
        \end{subfigure}
        \begin{subfigure}[t]{0.2\textwidth}
            \includegraphics[width=\linewidth]{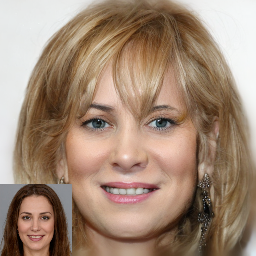}
            \centering
            \vspace{2pt}
            \includegraphics[width=\linewidth]{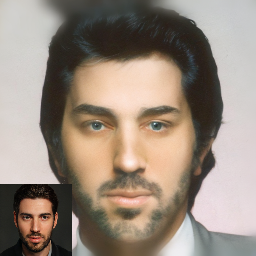}
            \centering
            \vspace{2pt}
            \fontsize{14pt}{16pt}\selectfont{\textbf{Ours}}
        \end{subfigure}
    \end{tabular}}
    \caption{Comparison of our method with vanilla and text-guided inpainting. Text prompts: \textit{smiling face with open mouth} (row 1) and \textit{add a long beard} (row 2).}
    \label{fig_3}
\end{figure}

\textbf{Adaptive Attribute-Guidance Module.} The vanilla denoising process in SD using the inpainting procedure can generate anonymized images of acceptable quality, but lacks attribute control. Even with textual guidance, control remains coarse and fails to provide fine-grained adjustments (see Fig. \ref{fig_3}). The vanilla denoising process in SD can be formulated as:
\begin{equation}
    \label{eq3}
    z_{t-1}=\sqrt{\bar{\alpha}_{t-1}} \widetilde{z}_0 + \sqrt{1 - \bar{\alpha}_{t-1} - \sigma^2_t} \epsilon_{\theta}(z_{t}, t, c) + \sigma_t\epsilon,
\end{equation}
where $\epsilon_{\theta}\left(z_{t}, t, c\right)$ shows the U-Net model, $\sigma_t$ means the standard deviation, and $\widetilde{z}_0$ is estimated from $z_{t}$ as:
\begin{equation}
    \label{eq4}
    \widetilde{z}_0 = \frac{1}{\sqrt{\bar{\alpha}_{t}}}\left(z_{t}-\sqrt{1-\bar{\alpha}_{t}} \epsilon_\theta\right).
\end{equation}
To address the mentioned issue and enable controllable anonymization, we first propose an adaptive attribute-guidance module. This module leverages a feature-matching loss $\mathcal{L}_{att}$ that aligns the generated image with the target in FaRL’s \cite{FARL_CVPR_2022} feature space. Specifically, the loss is implemented as a mean squared error (MSE) between the features of the intermediate prediction $\widetilde{x} = \mathcal{D}(\widetilde{z}_0)$ and $x_{tgt}$. Its gradient with respect to $\widetilde{z}_0$ provides a corrective direction to adjust latent vectors toward the target attributes \cite{Diff-Privacy_TCS_2024}. A larger gradient indicates a greater mismatch between the generated and target features, prompting stronger corrections, while a smaller gradient reflects closer alignment and thus applies only subtle adjustments. The module is formulated as:
\begin{equation}
    \label{eq5}
    \mathcal{M}_t = w_t \nabla_{\widetilde{z}_0} \mathcal{L}_{att}\left(\widetilde{x}, x_{tgt}\right),
\end{equation}
where $w_t = \lambda \sigma_t$, with the guidance weight $\lambda$~controlling the trade-off between target attribute enforcement and inpainting capability. $\sigma_t$ is adaptively decreased during the denoising process, allowing for stronger guidance in the early noisy steps and a weaker influence as the image becomes cleaner. Then we modify the reverse denoising process as:
\begin{equation}
    \label{eq6}
    \hat{z}_{t-1}=z_{t-1} - \mathcal{M}_t.
\end{equation}
Finally, the anonymized face image $x_a$ is generated by decoding the latent representation, $x_a = \mathcal{D}(\hat{z}_0)$. Additional details of the proposed method are provided in Alg. \ref{alg1}.

\textbf{Localized Anonymization.} As shown in Fig. \ref{fig_2}, our framework supports selective masking of specific regions (e.g., leaving the eyes or mouth unmasked $M'\notin\{eyes, lips\}$) instead of masking the entire face. The resulting masked image $x_m = x \odot (1 - M')$ is then constructed as before and used as input to the model, enabling localized anonymization while preserving critical facial details. This capability allows for a more adaptable trade-off between privacy protection and practical usability. It is particularly valuable in applications such as medical imaging, where patient privacy must be protected while retaining clinically relevant information.\par

\begin{algorithm}[t]
  \caption{Proposed method algorithm}
  \label{alg1}
  \textbf{Input:} real image $x$, synthesized target image $x_{tgt}$, guidance weight $\lambda$, and number of denoising steps $T'$ \\
  \textbf{Output:} anonymized image $x_a$
  \begin{algorithmic}[0]
  \State $z_0=\mathcal{E}(x)$, ${z}_T=\sqrt{\bar{\alpha}_T} z_0+\sqrt{1-\bar{\alpha}_T} \epsilon, \epsilon \sim \mathcal{N}(0,1)$
  \vspace{0.2em}
  \State $M = f(x)$, $x_m = x \odot (1 - M)$, and $c = \mathcal{E}(x_m)$ 
  \vspace{0.2em}
  \For{\textbf{all} $t$ from $T'$ to 1}
       \vspace{0.2em}
       \State $\widetilde{z}_0 = \frac{1}{\sqrt{\bar{\alpha}_{t}}}\left(z_{t}-\sqrt{1-\bar{\alpha}_{t}} \epsilon_\theta\right)$
       \vspace{0.2em}
       \State $z_{t-1}=\sqrt{\bar{\alpha}_{t-1}} \widetilde{z}_0 + \sqrt{1 - \bar{\alpha}_{t-1} - \sigma^2_t} \epsilon_{\theta}(z_{t}, t, c) + \sigma_t\epsilon$
       \vspace{0.6em}
       \State $\mathcal{M}_t = w_t \nabla_{\widetilde{z}_0} \mathcal{L}_{att}\left(\widetilde{x} =\mathcal{D}(\widetilde{z}_0), x_{tgt}\right)$
       \vspace{0.4em}
       \State $\hat{z}_{t-1}=z_{t-1} - \mathcal{M}_t$
       \vspace{0.2em}
  \EndFor
  \State return $x_a = \mathcal{D}(\hat{z}_0)$
  \end{algorithmic}
\end{algorithm}

\section{Experiments}
\label{sec:exp}
\subsection{Experimental Setting}
\textbf{Datasets and Benchmarks.} Following FAMS \cite{FAMS_WACV_2025}, we conduct experiments on the CelebA-HQ \cite{CelebA-HQ_2017_arXiv} and FFHQ \cite{FFHQ_CVPR_2019} datasets. From each dataset, we select a subset of 1,000 images, each representing a distinct identity. The selected images are divided into two groups based on gender (male and female), and anonymization is performed by using a target image of the same gender for each group. Target identities are chosen from the “This Person Does Not Exist” database$^{2}$. We utilize one GAN-based (FALCO) \cite{FALCO_CVPR_2023} and two state-of-the-art (SOTA) diffusion-based anonymization methods (CAMOUFLaGE \cite{CAMOUFLaGE_arXiv_2024} and FAMS \cite{FAMS_WACV_2025}) as benchmarks.\par

\textbf{Evaluation Metrics.} We evaluate the effectiveness of anonymization using the re-identification rate (Re-ID) \cite{FAMS_WACV_2025}. The same input and its anonymized counterpart are aligned and cropped with MTCNN \cite{MTCNN_2016_IEEE-SP}, then passed through FaceNet \cite{Facenet_2015_CVPR} to extract identity vectors. Re-ID is computed by measuring the cosine similarity between these vectors. Additionally, we report the Fréchet inception distance (FID) \cite{FID_2017_ANIPS}, visual DNA distance \cite{V-DNA_CVPR_2023}, and structural similarity index (SSIM) \cite{SSIM_2004_IEEE-IP} to evaluate the visual quality of the anonymized images.\par

\textbf{Implementation Details.} Our implementation is based on the Stable Diffusion model \cite{Stable_diff_2022_CVPR}, using $T = 1000$ steps for forward diffusion to obtain noisy images and $T' = 45$ steps for reverse diffusion to generate anonymized images, with a guidance weight of $\lambda = 0.8$. For masking, we utilize a face parsing model \cite{EHANet_MDPI_2020} that segments a face into 19 semantic regions, including skin, eyes, and other facial features.\par

\let\thefootnote\relax\footnotetext{$^{2}$ \href{https://thispersonnotexist.org/}{https://thispersonnotexist.org/}}

\subsection{Comparison Study}
\textbf{Assessment of Anonymization Ability.} The quantitative results in the second column of Tab. \ref{tab_1} show improvement in the Re-ID rate compared to the SOTA GAN-based and diffusion-based methods. Since we use target images to guide the anonymization process, for a fair comparison, we select four male and four female target images. Male targets are used to anonymize male data, and female targets for female data across both datasets. The final result is obtained by averaging the anonymized images generated with all selected targets. \par
It is worth noting that previous methods rely on training generative models to generate anonymized samples, which limits their performance to the diversity of the training data. In contrast, our approach requires no additional training and can generalize effectively to unseen data. Given an input and target image, our method generates an image in $\approx$10 seconds, faster than FAMS \cite{FAMS_WACV_2025} ($\approx$30 seconds) due to its high inference steps ($T' = 200$). FALCO \cite{FALCO_CVPR_2023} and CAMOUFLaGE \cite{CAMOUFLaGE_arXiv_2024} are near real-time but require training on new datasets. All experiments are conducted on a single NVIDIA RTX 4090.\par

\begin{figure}[t]
    \centering
    \resizebox{\columnwidth}{!}
    {\begin{tabular}{cccc}
        \begin{subfigure}[t]{0.2\textwidth}
            \includegraphics[width=\linewidth]{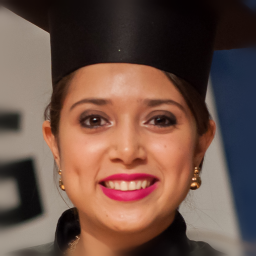}
            \centering
            \vspace{2pt}
            \includegraphics[width=\linewidth]{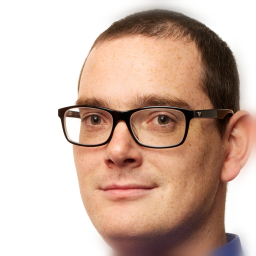}
            \centering
            \vspace{2pt}
            \includegraphics[width=\linewidth]{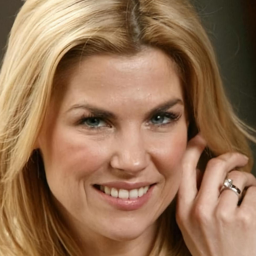}
            \centering
            \vspace{2pt}
            \fontsize{14pt}{16pt}\selectfont{Original Images}
        \end{subfigure}
        \begin{subfigure}[t]{0.2\textwidth}
            \includegraphics[width=\linewidth]{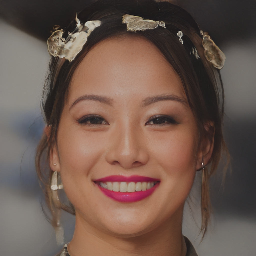}
            \centering
            \vspace{2pt}
            \includegraphics[width=\linewidth]{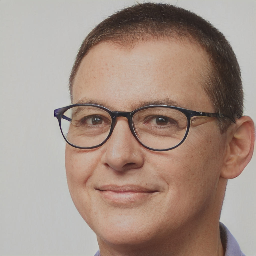}
            \centering
            \vspace{2pt}
            \includegraphics[width=\linewidth]{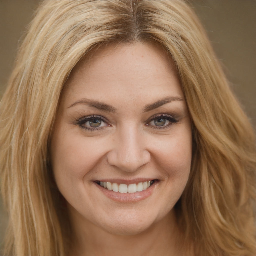}
            \centering
            \vspace{2pt}
            \fontsize{14pt}{16pt}\selectfont{FALCO \cite{FALCO_CVPR_2023}}
        \end{subfigure}
        \begin{subfigure}[t]{0.2\textwidth}
            \includegraphics[width=\linewidth]{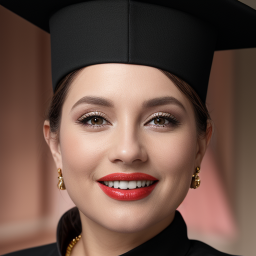}
            \centering
            \vspace{2pt}
            \includegraphics[width=\linewidth]{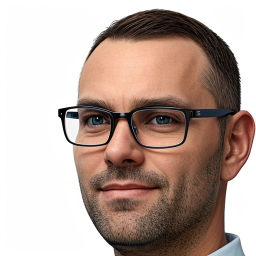}
            \centering
            \vspace{2pt}
            \includegraphics[width=\linewidth]{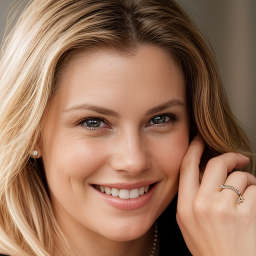}
            \centering
            \vspace{2pt}
            \fontsize{12pt}{14pt}\selectfont{CAMOUFLaGE \cite{CAMOUFLaGE_arXiv_2024}}
        \end{subfigure}
        \begin{subfigure}[t]{0.2\textwidth}
            \includegraphics[width=\linewidth]{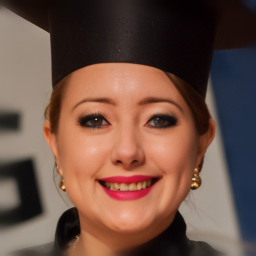}
            \centering
            \vspace{2pt}
            \includegraphics[width=\linewidth]{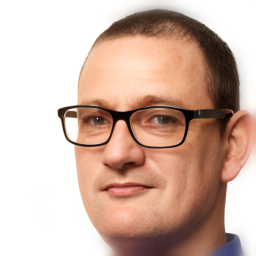}
            \centering
            \vspace{2pt}
            \includegraphics[width=\linewidth]{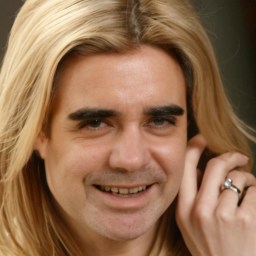}
            \centering
            \vspace{2pt}
            \fontsize{14pt}{16pt}\selectfont{FAMS \cite{FAMS_WACV_2025}}
        \end{subfigure}
        \begin{subfigure}[t]{0.2\textwidth}
            \includegraphics[width=\linewidth]{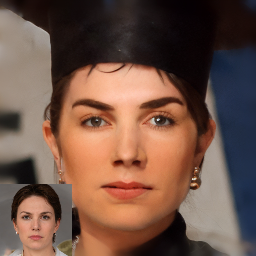}
            \centering
            \vspace{2pt}
            \includegraphics[width=\linewidth]{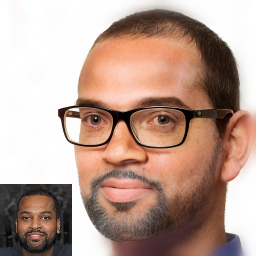}
            \centering
            \vspace{2pt}
            \includegraphics[width=\linewidth]{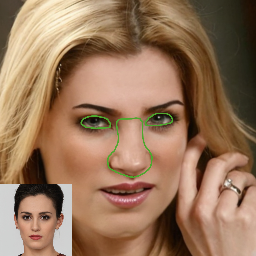}
            \centering
            \vspace{2pt}
            \fontsize{14pt}{16pt}\selectfont{\textbf{Ours}}
        \end{subfigure}
    \end{tabular}}
    \caption{Comparison of visual quality between our method and recent anonymization methods on both datasets.}
    \label{fig_4}
\end{figure}

\textbf{Assessment of Image Quality.} We conduct both quantitative and qualitative evaluations of the generated images. Columns 3–5 in Tab. \ref{tab_1} report the quantitative results. Although our method has a lower SSIM than FAMS \cite{FAMS_WACV_2025}, it achieves the best FID and Visual-DNA scores, showing that the anonymized images look more natural. Qualitative results are shown in Fig. \ref{fig_4}. FALCO \cite{FALCO_CVPR_2023} alters image structure, CAMOUFLaGE \cite{CAMOUFLaGE_arXiv_2024} modifies the background, and FAMS \cite{FAMS_WACV_2025} sometimes changes gender. These methods try to preserve facial attributes but lack control, whereas ours enables fine-grained attribute editing via target images while keeping the background through masking.\par

\begin{table}[t]
\centering
\resizebox{\columnwidth}{!}
{\begin{tabular}{c|cc|c|c|c}
\hline
\multirow{2}{*}{\textbf{Method}} & \multicolumn{2}{c|}{\textbf{Re-ID rate}$(\downarrow)$} & \multirow{2}{*}{\textbf{FID}$(\downarrow)$} & \multirow{2}{*}{\textbf{V-DNA}$(\downarrow)$} & \multirow{2}{*}{\textbf{SSIM}$(\uparrow)$} \\ \cline{2-3} & \multicolumn{1}{l}{CelebA-HQ} & \multicolumn{1}{l|}{FFHQ} & & & \\ \hline
FALCO & \underline{0.045} & \underline{0.069} & 45.256 & 6.221 & 62.79 \\
CAMOUFLaGE & 0.204 & 0.238 & 39.585 & 7.683 & 51.27 \\
FAMS & 0.138 & 0.299 & \underline{25.404} & \underline{4.307} & \textbf{79.12} \\
\textbf{Ours} & \textbf{0.015} & \textbf{0.047} & \textbf{23.193} & \textbf{3.794} & \underline{76.40} \\ \hline                       
\end{tabular}}
\caption{Quantitative assessments of anonymization and image quality. For the quality metrics, we report the average results across both CelebA-HQ and FFHQ datasets. The best and second-best results are marked in bold and underlined.}
\label{tab_1}
\end{table}

\subsection{Ablation Studies}
\textbf{Impact of Localized Anonymization.} As shown in Tab. \ref{tab_2}, localized anonymization increases the Re-ID rate compared to full-face masking, since some identity cues are preserved. Retaining small areas, such as the lips or eyes, causes only a slight increase, while preserving more distinctive regions, like the nose or a combination of regions, results in higher Re-ID scores. However, the scores remain low and are still comparable to or better than those of existing anonymization methods. This demonstrates that our approach can balance identity anonymization and utility by selectively retaining clinically relevant facial regions.\par

\begin{figure}[h]
    \centering
    \resizebox{\columnwidth}{!}
    {\begin{tabular}{ccccc}
        \begin{subfigure}[t]{0.2\textwidth}
            \includegraphics[width=\linewidth]{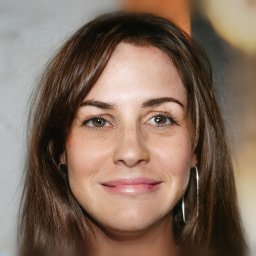}
            \centering
            \vspace{2pt}
            \fontsize{17pt}{19pt}\selectfont{\textbf{Ours}}
        \end{subfigure}
        \begin{subfigure}[t]{0.2\textwidth}
            \includegraphics[width=\linewidth]{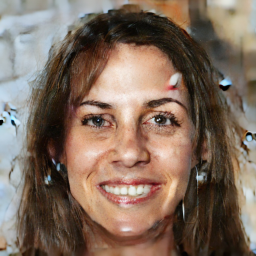}
            \centering
            \vspace{2pt}
            \fontsize{17pt}{19pt}\selectfont{w/o adaptive}
        \end{subfigure}
        \begin{subfigure}[t]{0.2\textwidth}
            \includegraphics[width=\linewidth]{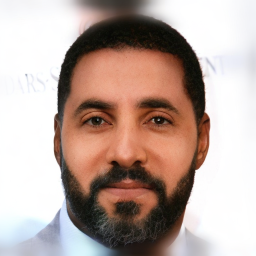}
            \centering
            \vspace{2pt}
            \fontsize{17pt}{19pt}\selectfont{\textbf{Ours}}
        \end{subfigure}
        \begin{subfigure}[t]{0.2\textwidth}
            \includegraphics[width=\linewidth]{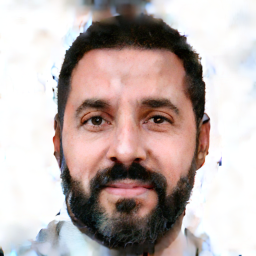}
            \centering
            \vspace{2pt}
            \fontsize{17pt}{19pt}\selectfont{w/o adaptive}
        \end{subfigure}
        \begin{subfigure}[t]{0.2\textwidth}
            \includegraphics[width=\linewidth]{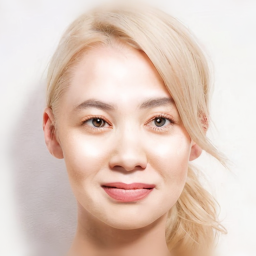}
            \centering
            \vspace{2pt}
            \fontsize{17pt}{19pt}\selectfont{\textbf{Ours}}
        \end{subfigure}
        \begin{subfigure}[t]{0.2\textwidth}
            \includegraphics[width=\linewidth]{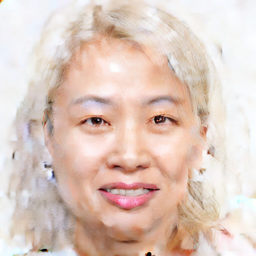}
            \centering
            \vspace{2pt}
            \fontsize{17pt}{19pt}\selectfont{w/o adaptive}
        \end{subfigure}
    \end{tabular}}
    \caption{Qualitative evaluation of the impact of removing $\sigma_t$}
    \label{fig_5}
\end{figure}
\textbf{Impact of Adaptive Weight Component $w_t$.} We also conduct an ablation study by removing the standard deviation $\sigma_t$ from $w_t$. As shown in Fig. \ref{fig_5}, this results in noticeable degradation in the quality of the generated images. The adaptive weight plays a key role: in the early steps, when the latent is highly noisy, strong guidance is needed to push generation toward the desired attributes. In the later steps, however, the image is much cleaner, and applying the same level of guidance would distort fine details or introduce artifacts. By gradually decreasing $\sigma_t$, the model balances these phases, ensuring effective attribute control early while preventing over-correction in later steps.\par

\begin{table}[t]
\centering
\resizebox{\columnwidth}{!}
{\begin{tabular}{ccccccc}
\cline{2-7}
\multirow{2}{*}{\textbf{}} & \multicolumn{6}{c}{\textbf{Re-ID rate}} \\ \cline{2-7} 
 & full-face & r. lips & r. eyes & r. eyebrows & r. nose & \begin{tabular}[c]{@{}c@{}}r. eyes, nose, \\ lips\end{tabular} \\ \hline
\multicolumn{1}{c|}{\begin{tabular}[c]{@{}c@{}}CelebA male \\ images\end{tabular}} & 0.0053 & 0.0211 & 0.0368 & 0.0895 & 0.0974  & 0.2421 \\ \hline
\end{tabular}}
\caption{Quantitative comparison of full-face versus localized anonymization. The full-face mask anonymizes the entire face, while localized masks reveal specific regions such as eyes, nose, etc.}
\label{tab_2}
\end{table}

\section{Conclusion}
\label{sec:conc}
In this paper, we introduced a diffusion-based framework for controllable and localized face anonymization. By leveraging the inpainting capabilities of latent diffusion models and incorporating adaptive attribute guidance, our method enables fine-grained control over facial attributes while preserving backgrounds and usability. Experiments on CelebA-HQ and FFHQ demonstrate that our approach achieves superior anonymization effectiveness and image quality compared to cutting-edge methods, without requiring additional training.\par

However, the framework relies on a target image to guide attribute control, which may limit its practicality in some scenarios. In future work, we plan to explore text-based or disentangled latent controls as alternatives, aiming to remove the dependency on explicit target images while retaining fine-grained controllability.

\vfill\pagebreak

\textbf{Acknowledgements.} {This work was supported by the Research Council of Finland (former Academy of Finland) Academy Professor project EmotionAI (grants 336116, 345122, 359894), HPC project FaceCanvas (grant number 364905), the University of Oulu \& Research Council of Finland Profi 7 (grant 352788), EU HORIZON-MSCA-SE-2022 project ACMod (grant 101130271), Academy Research Fellow project (grant 355095), and Infotech Oulu.}

\bibliographystyle{IEEEbib}
\bibliography{main}

\end{document}